\documentclass[pdflatex,sn-basic,Numbered]{sn-jnl}

\usepackage{graphicx}%
\usepackage{amsmath,amssymb,amsfonts}%
\usepackage{amsthm}%
\usepackage{mathrsfs}%
\usepackage[title]{appendix}%
\usepackage{xcolor}%
\usepackage{textcomp}%
\usepackage{manyfoot}%
\usepackage{booktabs}%
\usepackage{algorithm}%
\usepackage{algorithmicx}%
\usepackage{algpseudocode}%
\usepackage{listings}%
\usepackage[english]{babel}
\usepackage[utf8]{inputenc}
\usepackage{comment}
\usepackage{tabularx}
\usepackage{makecell}
\usepackage{multicol, multirow}
\usepackage{todonotes}
\usepackage{enumitem}
\usepackage{hyperref}
\usepackage{float}
\usepackage{url}
\usepackage{hyperref}

\begin{document}

\title[DELICATE]{DELICATE: Diachronic Entity LInking using Classes And Temporal Evidence}


\author*[1,3]{\fnm{Cristian} \sur{Santini}}\email{c.santini12@unimc.it}
\author[2]{\fnm{Sebastian} \sur{Barzaghi}}\email{sebastian.barzaghi2@unibo.it}
\author[1]{\fnm{Paolo} \sur{Sernani}}\email{paolo.sernani@unimc.it}
\author[1]{\fnm{Emanuele} \sur{Frontoni}}\email{emanuele.frontoni@unimc.it}
\author[3]{\fnm{Mehwish} \sur{Alam}}\email{mehwish.alam@telecom-paris.fr}

\affil*[1]{\orgname{University of Macerata}, \orgaddress{\city{Macerata}, \country{Italy}}}
\affil[2]{\orgname{University of Bologna}, \orgaddress{\city{Bologna}, \country{Italy}}}
\affil[3]{\orgname{Télécom Paris, Institut Polytechnique de Paris}, \orgaddress{\city{Palaiseau}, \country{France}}}


\abstract{In spite of the remarkable advancements in the field of Natural Language Processing, the task of Entity Linking (EL) remains challenging in the field of humanities due to complex document typologies, lack of domain-specific datasets and models, and long-tail entities, i.e., entities under-represented in Knowledge Bases (KBs). The goal of this paper is to address these issues with two main contributions. The first contribution is DELICATE, a novel supervised method for EL on historical Italian which combines a BERT-based encoder with contextual information from Wikidata to select appropriate KB entities using temporal plausibility and entity type consistency. The second contribution is ENEIDE, a multi-domain EL corpus in historical Italian semi-automatically extracted from two annotated editions spanning from the 19th to the 20th century and including literary and political texts. Results show how DELICATE outperforms other EL models on ENEIDE even if compared with generative approaches using Large Language Models (LLMs) for context augmentation and candidate selection. Moreover, further analyses reveal how DELICATE confidence scores and features sensitivity provide results which are more explainable and interpretable than purely neural methods.}


\maketitle

\section{Introduction}
\label{sec:introduction}

\begin{figure*}
    \centering
    \includegraphics[width=1\textwidth]{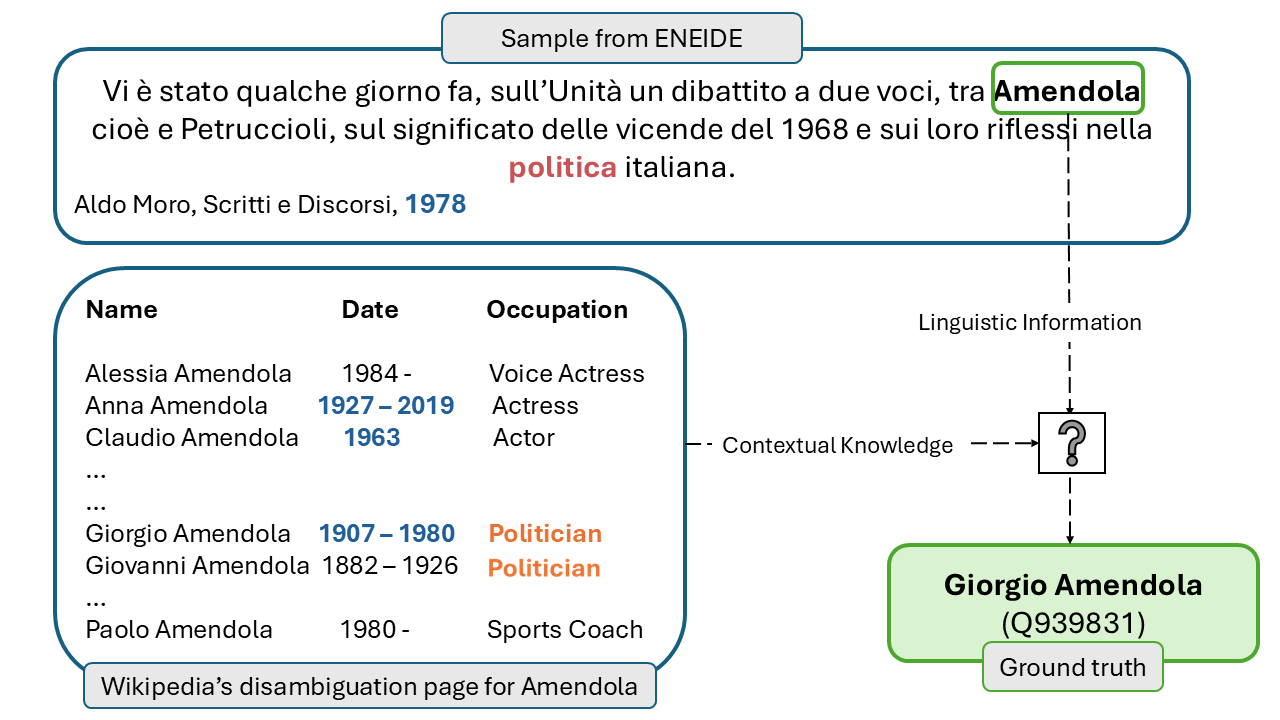}
    \caption{Representation of a sample sentence from the ENEIDE dataset containing a named entity and candidate entities from Wikipedia presented in a disambiguation page. The image shows how contextual information provided in external KBs allows to better determine the correct entity referenced in a historical text.}
    \label{fig:example}
\end{figure*}

Entity Linking (EL) is the task of identifying and disambiguating entities mentioned in a text by linking them to their corresponding entries in structured Knowledge Bases (KBs) such as Wikipedia, Wikidata\footnote{\href{https://www.wikidata.org}{www.wikidata.org}}~\citep{vrandevcic2014wikidata}, etc. EL is a key component of many Knowledge Extraction (KE) applications in the Digital Humanities (DH), where it enables researchers to trace references to people, locations, organizations, and other named entities across historical sources. Despite the wide application of EL in various domains, general-purpose EL models often perform poorly on historical documents. These texts pose specific challenges, such as linguistic variations over time and the need for accurate chronological context, which are typically overlooked by State-of-The-Art (SoTA) models that rely solely on linguistic information~\citep{wu2020scalable, de2022multilingual, limkonchotiwat-etal-2023-mrefined}.
Figure~\ref{fig:example} shows a sentence from a text written in 1978 by Aldo Moro, an Italian politician, containing a reference regarding a person mentioned with the surname ``Amendola". As the figure shows, there are multiple politicians in Wikipedia with that surname, therefore the information given only in the text is not enough to find a correct candidate. As this example proves, temporal information given in Wikidata about the entity's birth date can guide EL models to a correct decision.

To address this issue, several domain-specific methods have been proposed~\citep{labusch_named_2020, linhares2022melhissa,graciotti2025musical} to incorporate contextual features such as entity types and temporal information derived from KBs for improving disambiguation accuracy. However, a common limitation of these existing approaches is their reliance on manually crafted rules, which can be rigid and may lack generalizability. In contrast, this paper proposes a method that automatically learns the relevance of contextual information, such as class equivalence and temporal distance, by using them as features in a supervised tree-ensemble model called Gradient Boosted Trees (GBTs). This traditional Machine Learning (ML) approach, compared to more complex Deep Neural Networks (DNNs), has the advantage of efficiently capturing statistical patterns based on numerical features without requiring extensive pre-training, large datasets, and powerful computational resources~\citep{grinsztajn2022tree}. Moreover, a further advantage of GBTs is to have a higher level of explainability compared to DNNs since each tree creates explicit if-then rules with clear decision boundaries based on feature thresholds  and the relevance of each feature can be measured based on permutation importance or gain-based importance~\citep{yasodhara2021trustworthiness}.

Moreover, the lack of benchmark datasets for EL in the DH is a pivotal issue that must be addressed for advancing NLP techniques in this field~\citep{survey_hist_ner_2023}. To the best of our knowledge, currently there are no datasets in Italian which allow to train historical EL models. Due to the increasing amount of data produced in the DH domain including corpora~\citep{ehrmann2022overview}, digital editions~\citep{cristofaro2025implementing}, and KBs~\citep{carriero2019arco}, it is crucial to understand how to combine these resources to advance the SoTA of EL for under-explored languages such as historical Italian. In this regard, Scholarly Digital Editions (SDEs)~\citep{sahle_what_2016} are a primary source of annotations for people, places, organizations, bibliographic resources, and other entities, providing a large number of samples of entities referenced in historical texts and disambiguated using Wikidata or other domain-agnostic KBs. As a consequence, this study aims to address the following Research Questions (RQs):

\begin{itemize}
    \item \textbf{RQ1}: How can we address the problem of EL in historical documents by combining approaches based on language models with tree-based methods using background knowledge, i.e. dates and classes, from Wikidata?
    \item \textbf{RQ2}: Can we leverage the information already present in SDEs to build a novel training and evaluation corpus for EL in Italian based on historical documents?
    \item \textbf{RQ3}: Can we analyse the features and scores of GBTs through statistical techniques to provide a clearer explanation of the behaviour of a supervised EL model in the historical domain?
\end{itemize}

To answer these questions, this study introduces DELICATE, a novel supervised architecture for EL designed specifically for historical texts. DELICATE combines the strengths of neural language models and tree-ensemble methods by integrating a bi-encoder model~\citep{pozzi2023named} based on BERT~\citep{devlin_bert_2019} for candidate retrieval with a supervised GBTs classifier for candidate re-ranking. The novel aspect of  DELICATE is its ability to incorporate class and temporal information from Wikidata as explicit features in the re-ranking process. In particular, DELICATE uses a GBTs classifier to re-rank the bi-encoder results by computing pairwise similarity scores based on features such as temporal distance, type compatibility, string similarity, and embedding distance. This allows the model to dynamically assess the contextual relevance of each candidate entity, which is important when dealing with diachronic variations in language and meaning. In addition, DELICATE re-ranker can be coupled with a Large Language Model (LLM) to enhance the classifier's decisions.

Our tests on historical Italian show how DELICATE achieves better performances on multiple benchmarks if compared with auto-regressive multilingual models~\citep{de2022multilingual} and even with SoTA approaches using LLMs for context augmentation and candidate selection. While we cannot directly claim that our architecture pairing a bi-encoder with a GBTs model can outperform systems using LLMs with huge number of parameters such as GPT-4o mini or LLaMa3-70B, our findings show that bi-encoder models paired with supervised re-rankers that capture contextual information from Wikidata have two benefits:

\begin{itemize}
\item They efficiently improve the candidate retrieval step based on temporal plausibility and typological consistency with a given mention using a lightweight and fast architecture;
\item They enhance the interpretability of EL results by allowing the measurement of the sensitivity of linguistic, temporal, and typological features during re-ranking and by providing more interpretable confidence scores.
\end{itemize}

As a consequence, DELICATE aims to be an approach that complements, rather than competes with, current EL architectures using LLMs~\citep{ding2024chatel, wang2025aelc, xin2025llmael}.

Given that there are currently no benchmark datasets in Italian with chronological information, we also introduce ENEIDE (\textit{Extracted Named Entities from Italian Digital Editions}), a publicly available EL corpus which provides a training and evaluation benchmark with texts in historical Italian spanning multiple domains and years. ENEIDE is semi-automatically constructed using annotations from two SDEs: Digital Zibaldone\footnote{\href{https://digitalzibaldone.net/}{digitalzibaldone.net}}~\citep{stoyanova_working_2023} and Aldo Moro Digitale\footnote{\href{https://aldomorodigitale.unibo.it/}{aldomorodigitale.unibo.it}}~\citep{barzaghi_amd_2025}, which comprise literary and political texts written between the 19th to the 20th century. By leveraging existing entity annotations in SDEs and linking them to Wikidata, ENEIDE provides a challenging benchmark for training and evaluating EL systems in Italian.

This paper is structured as follows. Section~\ref{sec:related_work} gives an overview of the current challenges of EL in the domain of Cultural Heritage (CH) and provides a brief analysis of EL corpora and neural models for EL in the humanities. Section~\ref{sec:methodology} contains the description of the architecture of DELICATE. Section~\ref{sec:delicate_llm} discusses how LLMs can be used to refine EL results on historical texts using our retrieval and re-ranking approach. Section~\ref{sec:dataset} presents ENEIDE and discusses the data preprocessing and annotation strategies employed for dataset creation. Section~\ref{sec:experiments} discusses the experimental results of DELICATE for the tasks of Entity Disambiguation (ED) and end-to-end EL on two historical Italian corpus: ENEIDE and MHERCL-ITA~\citep{graciotti2025musical}. Section~\ref{sec:discussion} gives an overview of the advantages and limitations of the approach herein presented and concludes the work.

\section{Related Work}
\label{sec:related_work}

This section details the related studies. First, Section \ref{sec:challenges} presents the foundations of the field of EL in the historical domain and the related work for historical Italian. Then, Section \ref{sec:approaches} discusses the SoTA with respect to neural EL approaches for historical texts. Finally, Section \ref{sec:el_and_llms} lists some of the most recent contributions investigating the use of LLMs for EL.

\subsection{Challenges of EL for Historical Italian}
\label{sec:challenges}

In the existing literature, EL is divided into two different tasks. The first is NER, in which the entity mentions are identified in text and classified according to different categories such as person, location, organization, or miscellaneous. The subsequent task is ED, in which detected mentions are linked to the respective identifier in a structured KB, such as Wikidata. EL approaches often rely on a separate NER component trained to identify surface forms of named entities within a text~\citep{labusch_named_2020}; however, several approaches have investigated the possibility of training NER and ED jointly within a single end-to-end architecture~\citep{kolitsas-etal-2018-end,limkonchotiwat-etal-2023-mrefined}. 

Despite the remarkable performances that the NER and ED systems have achieved in contemporary web-crawled data~\citep{de2022multilingual, limkonchotiwat-etal-2023-mrefined}, they under-perform when applied to humanistic documents (such as historical press articles, books, literary texts, or letters), which may be excluded from the training corpora and therefore are not observed in the learning phase. In a recent survey~\citep{survey_hist_ner_2023}, researchers highlight some of the most important contributions in the field of EL with historical documents and highlight the main challenges encountered. The four main challenges include: (i) the variety of documents (newspapers, letters, memoirs, books); (ii) the presence of noise in the input data due to data processing techniques, such as Optical Character Recognition (OCR); (iii) the problem of linguistic and chronological variations; and (iv) the lack of standardized benchmarks across multiple languages and genres. 

Due to the need to adapt the NER and EL approaches for documents provided by Cultural Heritage (CH) institutions, several resources have been created in order to provide standardized benchmarks in the field of historical EL. With respect to Italian, a pivotal work has been carried out in~\citet{paccosi-palmero-aprosio-2022-kind}, where the authors describe \textit{KIND}, a multi-domain dataset for NER extracted from several typologies of texts, including news, literary texts, and political works. Similar to ENEIDE, this dataset includes texts from the political domain, extracted from the De Gasperi Corpus~\citep{sprugnoli2016fifty} and Aldo Moro Digitale~\citep{barzaghi_amd_2025}. However, this work does not contain links to a KB for disambiguation.

With respect to the literary domain, one of the first examples of datasets in Italian containing annotated references to literary works (such as citations to books, monographs, and essays), is \textit{LinkedBooks}~\citep{colavizza_annotated_2017}. In spite of the contribution of this resource for NER with literary entities, the dataset does not include more general entities (such as person, location, or organization) and the references to works are not disambiguated using Wikidata identifiers. More recently, in~\citet{ajmc_2024}, the authors present a manually annotated dataset of named entity references in classical commentaries of Sophocles' \textit{Ajax}. The dataset, AJMC, was included in the HIPE-2022 shared task~\citep{ehrmann2022overview}. It was used for training and evaluating the systems for the detection of references to both general entities (person, location, and date) and domain-specific entities (primary and secondary literary sources), as well as for the disambiguation of the entities by using Wikidata as an external KB in English, French and German. Similar to the literary texts in ENEIDE, entities are often referenced through abbreviations, as in \textit{Cic.} for \textit{Cicero} (a common practice in philological texts). Another feature that AJMC shares with ENEIDE is the fact that it contains several annotated references to literary works, either disambiguated with Wikidata or labeled as NIL if not present in the KB. 

A very recent attempt to provide a benchmark for EL in Italian was presented in~\citet{graciotti2025ke}. In this paper, the authors present MHERCL-ITA\footnote{\href{https://github.com/polifonia-project/KE-MHISTO/blob/main/Datasets/MHERCL\_ITA.json}{github.com/polifonia-project/KE-MHISTO/blob/main/Datasets/MHERCL\_ITA.json}}, the first gold standard for EL in historical Italian, containing texts extracted from music periodicals published between 1853 to 1943. In total, MHERCL-ITA contains 533 sentences with 2,431 manual annotations of entities linked to Wikidata. In spite of its relevance for historical Italian, the main limitation of MHERCL-ITA is its small scale, since it only comprises a test set and is more suitable to evaluate pre-trained EL models or LLMs for zero-shot EL.

\subsection{Neural Approaches for Historical EL}
\label{sec:approaches}

One of the seminal works on neural end-to-end EL applied to historical documents has been presented in~\citet{linhares2022melhissa}. This system employs several strategies, including an OCR correction mechanism, a probabilistic entity table map, a multilingual KB that contains entity representations obtained from the multilingual Wikipedia and a BiLSTM network which is trained to compute the matching probability between the context in which an entity appears and its representation in the KB. Despite the remarkable performances shown on multilingual historical press articles (excluding Italian), the model makes limited use of contextual knowledge by simply applying rule-based filters on the results using temporal constraints, type-related information, and edit distance between entity labels and surface form. Instead, one of the advantages of our approach is the possibility to learn the relevance of contextual information by employing supervised learning in the candidate re-ranking step. 

The ED problem, which is the task of linking already detected mentions, is addressed for historical documents in~\citet{labusch_named_2020}. This work proposes a multilingual architecture for English, German, and French, based on three components: (a) an entity-candidate lookup; (b) an entity-candidate evaluation; and (c) an entity-candidate ranking. In (a), a word embedding model is used to perform a \textit{k-NN} search on a dense index of Wikipedia titles to find the most similar entities with respect to a query mention. In this step, the system also makes use of the date of publication of a document, provided in the metadata, to consider only the entities whose birth or inception are not posterior to the document creation as potential candidates for linking. In the entity-candidate evaluation, a sentence-matching algorithm based on a BERT encoder is used to compare the context of the mention with the Wikipedia text snippets of the candidates returned in (a). In the entity-candidate ranking step, the similarity scores of step (a) and step (b) are fed to a Random Forest algorithm which computes the matching probabilities between the items in the candidate set and the query mention. More recently,~\citet{graciotti2025musical} proposed the use of rule-based constraints on Wikidata to refine the candidate retrieval step of a bi-encoder architecture to perform ED on historical music periodicals in English. More concretely, they paired BLINK~\citep{wu2020scalable} with multiple rule-based constraints in which the entities retrieved are verified for compliance with respect to the date of publication of a document and the entity type given in the metadata.

One of the main limitations of these approaches is that using filters based on temporal and semantic information in the lookup step may cause the system to incorrectly exclude from linking correct candidates. In contrast, our solution for this problem consists in employing temporal and semantic information in the final re-ranking phase. In DELICATE, information related to dates and types of the top-\textit{k} candidates retrieved is extracted from Wikidata and is used as a feature to train a supervised classifier for the task of learning optimal similarity thresholds between input mentions and attributes of the candidates.

\subsection{EL with LLMs}
\label{sec:el_and_llms}

Recent advances in LLMs have motivated researchers to explore their application to the EL task. This section provides an overview of contemporary approaches that leverage LLMs for ED, discussing their methodologies and limitations in the context of historical text processing.

One of the first attempts to integrate LLMs with ED is presented in~\citet{lai2022improving}, where the authors propose an approach for performing ED directly on Wikidata. Their method employs an ElasticSearch index for candidate retrieval over Wikidata entities and their descriptions, followed by a sequence-to-sequence (seq2seq) model trained to generate entity descriptions for a given mention based on an input sentence. However, this approach presents two main limitations: (i) the seq2seq model is trained exclusively on English texts, limiting its applicability in multilingual settings; and (ii) the method does not incorporate temporal information from the KB to refine disambiguation results.

ChatEL~\citep{ding2024chatel} introduces a three-step prompting framework where LLMs generate auxiliary content and perform multi-choice entity selection from BLINK-generated candidates. Despite its innovative use of conversational prompting, ChatEL exhibits several limitations: (i) no contextual information from KBs is exploited during disambiguation; (ii) the contextual information generated by LLMs may contain hallucinations and be biased by implicit assumptions; and (iii) the reliance on proprietary LLMs (GPT-4) raises concerns regarding reproducibility and environmental costs.

LLMAEL~\citep{xin2025llmael} treats LLMs as ``context augmenters'' rather than direct ED executors, generating auxiliary content for entity mentions that are subsequently fed to specialised models such as ReFinED and BLINK. While this approach demonstrates the potential of LLMs for enriching contextual representations, it presents significant drawbacks: (i) generated mention descriptions may perpetuate biases inherent in LLMs when applied to historical texts; (ii) explainability issues arise from the fact that the context used by the disambiguation algorithm is generated by a model lacking interpretability; and (iii) to mitigate hallucination risks, the authors propose using large-scale LLMs with 70B parameters, a solution that is impractical for public CH institutions, which may not have the computational infrastructures of private tech companies.

AELC~\citep{wang2025aelc} combines context augmentation with candidate retrieval and re-ranking through a combination of LLMs, bi-encoders, and sentence encoders. The LLM is employed for both context augmentation and candidate selection. Similarly to ChatEL and LLMAEL, this architecture employs LLMs with 70B parameters multiple times for a single text passage, resulting in computational costs that are unsustainable for CH institutions such as archives and libraries, which often manage hundreds of thousands or even millions of records.

The integration of KBs with LLMs for ED has been explored in~\citet{pons2024kg}, where the authors leverage YAGO and DBpedia class hierarchies for iterative candidate pruning, demonstrating improvements exceeding 25 percentage points on out-of-domain KORE data. The intuition of using class hierarchies to prune candidates is conceptually similar to the entity type matching employed in DELICATE for candidate re-ranking. However, this architecture does not leverage temporal information contained in KBs, which is crucial for diachronic EL as discussed in Section~\ref{sec:introduction}.

DeepMEL~\citep{wang2026deepmel} introduces a multi-agent collaboration framework with specialised components for modal fusion, candidate adaptation, and cloze-style disambiguation. As a multimodal framework requiring images, it addresses fundamentally different challenges than text-only historical EL and is therefore not directly comparable to DELICATE.

Finally,~\citet{zhang2025lifelong} proposes a lifelong learning approach for ED that employs weight-based pruning for continual domain learning, where entity domains are learned sequentially. This method updates redundant parameters to learn new entity domains while retaining important parameters to preserve knowledge from previously learned domains. However, this approach has not been trained on the historical domain and does not incorporate temporal information during disambiguation.

While these works demonstrate that LLMs can effectively enhance the robustness of EL systems based on candidate retrieval, they present several limitations that DELICATE aims to address. First, although context augmentation can provide additional background information for the candidate selection step, it may introduce hallucinations due to the inherent biases of LLMs. Moreover, none of the surveyed approaches consider the provenance context of a text to provide time-sensitive context augmentation. Second, during candidate retrieval, DELICATE enriches the candidate set with temporal and categorical attributes from Wikidata. This information is crucial for re-ranking candidates based on temporal and ontological consistency with the in-text mention.

In contrast, all LLM-based methods except KB-enhanced approaches~\cite{pons2024kg} do not incorporate information from KBs during disambiguation. Third, all LLM-based approaches rely on models with more than 70B parameters or proprietary LLMs (e.g., ChatGPT) to mitigate hallucination risks and perform multiple forward passes for a single text passage (at least two for context augmentation and candidate selection). This computational overhead is impractical for CH institutions that often need to process millions or even billions of pages of text. DELICATE, which relies on a small BERT-based bi-encoder and a supervised GBT classifier, aims to provide an accurate disambiguation system that is not only computationally reproducible but also environmentally sustainable. Finally, it should be noted that none of the surveyed papers using LLMs engage with the specific use case of historical EL, which involves unique challenges such as OCR noise, linguistic variations, and long-tail entities that are under-represented in contemporary KBs.

\section{DELICATE}
\label{sec:methodology}

\textbf{Preliminaries:} The ED task aims to resolve the ambiguity of a given mention \( m \) appearing in a context of words \( c \) by linking it to the most appropriate entity \( e_i \) from a predefined finite set \( E = \{e_1, \dots, e_i, \dots, e_n\} \), where \( E \) is a collection of entities extracted from a KB and \( e_i \in E \).  

Each mention \( m \) is characterized by a set of attributes \( A_m = \{s_m, t_m, d_m\} \), where:  
\begin{itemize}
\item \( s_m \) is the surface form with which an entity is referred in the text,
\item \( t_m \) denotes the entity type associated with \( m \), given in the ground-truth annotation or by a NER model,
\item \( d_m \) represents the timestamp of the mention, inferred from the document date. 
\end{itemize}

Similarly, each entity \( e_i \) is described by a set of attributes \( A_{e_i} = \{s_{e_i}, t_{e_i}, d_{e_i}\} \), where:  
\begin{itemize}
\item \( s_{e_i} \) is the surface form of an entity in the KB, e.g., the Wikidata label in Italian, 
\item \( t_{e_i} \) denotes the entity type as defined in the KB,  
\item \( d_{e_i} \) is a date associated with \( e_i \) in the KB.
\end{itemize}

\begin{figure*}
    \centering
    \includegraphics[width=1\textwidth]{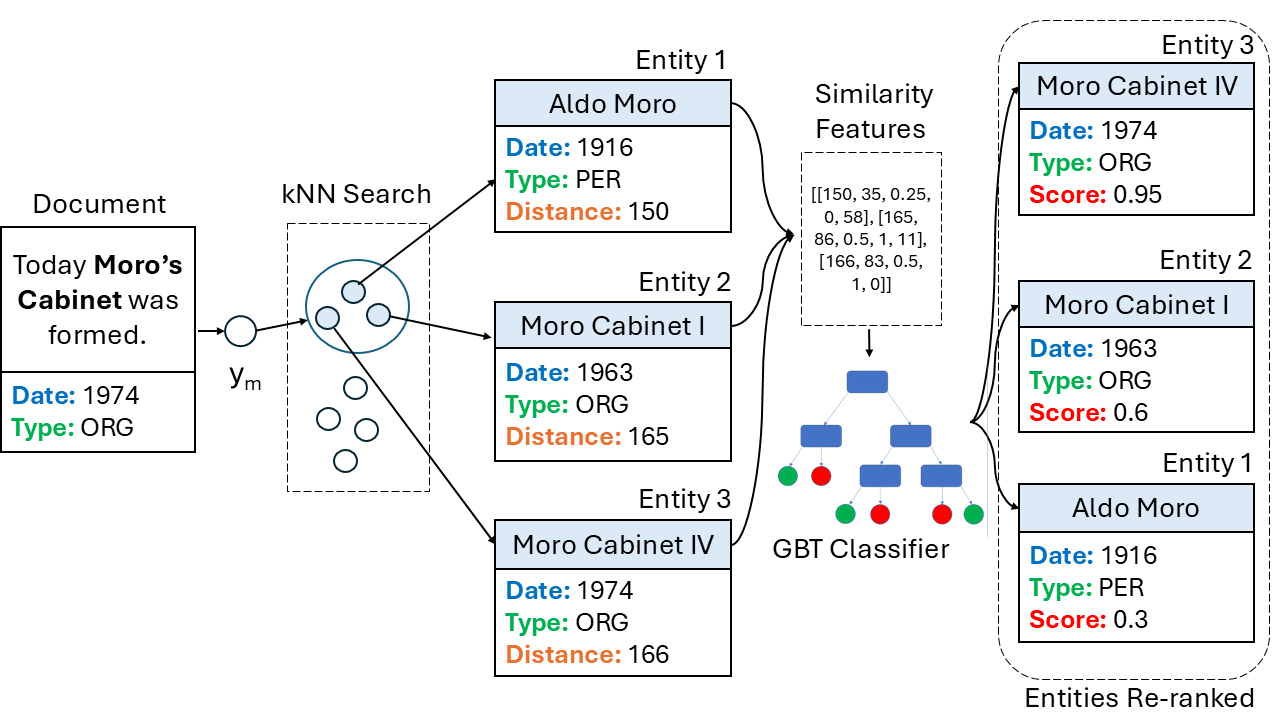}
    \caption{High-level representation of DELICATE. 
  At first, similar entities with respect to a given mention are retrieved from a dense index of Wikipedia entities by performing k-NN search using the BLINK bi-encoder.  In the second step, entities are re-ranked by a GBTs model which takes as input pairwise similarity features computed for each mention-entity pair.}
    \label{fig:elite}
\end{figure*}

\noindent \textbf{Architecture:} \textit{DELICATE} is a novel architecture to perform ED on digitized historical and humanistic texts which makes use of entity types and temporal information from Wikidata to disambiguate entities in diachronic and historical texts. Figure~\ref{fig:elite} shows the overall architecture of DELICATE, which relies on the following components: 
\begin{itemize}
    \item a \textbf{bi-encoder} based on the BLINK architecture~\citep{wu2020scalable} which encodes a mention and its surrounding context in a latent representations to be used for \textit{k}-NN search,
    \item a \textbf{dense vector index}, implemented with FAISS~\citep{johnson2019billion}, which contains embeddings for each entity in the Italian Wikipedia encoded from the first Wikipedia paragraph using the bi-encoder, 
    \item a \textbf{lookup table}, which contains structured information about the entities in Wikipedia, such as entity types and dates extracted from Wikidata,
    \item a \textbf{GBTs classifier}, which makes a pairwise comparison between the mention and each Wikipedia entity retrieved with k-NN search and computes a probability score for each candidate entity to predict the most plausible match.
\end{itemize}

While the use of a bi-encoder for candidate retrieval has already been explored in~\citet{wu2020scalable} for English and in~\citet{pozzi2023named} for Italian, the novelty of this work consists in refining the results of the bi-encoder in an additional re-ranking step. In this step, a supervised classifier trained on labeled data takes as input a set of similarity features between the entity mention and the Wikipedia candidates, such as temporal distance, type equivalence, string similarity, and $L^2$ embeddings distance, and returns a set of probability scores which are used to filter and re-rank candidates. The hypothesis of this work is that the use of carefully selected features to model $(m, e_i)$ similarity based on the respective set of attributes $A_m$ and $A_{e_i}$ allows to improve the performance of ED systems on historical datasets. 

\subsection{Bi-encoder and Dense Vector Index }

For performing candidate retrieval, DELICATE adopts a BLINK model originally proposed in~\citet{wu2020scalable}. BLINK is an ED model which adopts a Transformer-based model to encode representations of entity references and Wikipedia candidates in the same latent space. In order to do so, BLINK encodes each mention and its surrounding context (i.e., words preceding and following) into a vector using a BERT-based model. The same model is used to encode representations of each entity inside Wikipedia by modeling the concatenation of the Wikipedia page title and its first 10 sentences into an embedding. While the mention representations are computed at inference time, the entity representations are precomputed and stored into a dense vector index implemented with FAISS. In BLINK, the scoring function is simply the dot product between the mention vector $y_m$ and the entity vector $y_{e_i}$, and is computed as follows:
\begin{equation}
    s(m, e_i) = y_m \cdot y_{e_i} 
\end{equation}

The loss function used to train the bi-encoder aims to maximize the dot product of the correct entity, while minimizing the scores of the incorrect ones, and is specified as follows: 

\begin{equation}
    L(m, e_i) = -s(m, e_i) + \log \sum_{j=1}^{B} \exp(s(m, e_j))
\end{equation}

DELICATE stores embeddings of entities in a FAISS index comprising $\approx 1.5M$ entities from the Italian Wikipedia. To perform candidate retrieval, the embedding of the text mention is used to perform k-NN search on the index which returns top-\textit{k} candidates sorted by Euclidean distance (or $L^2$ distance). The implementation of DELICATE relies on the frozen weights of BLINK$_{ITA}$ released in~\citet{pozzi2023named}.

\subsection{Lookup Table}
\label{sec:lookup_table}

To obtain additional information about the candidates retrieved using the bi-encoder, DELICATE uses a lookup table stored in a SQLITE database to associate each entity in Wikipedia with structured information. First, each entity in the index is linked to its Wikidata identifier. Then, each Wikipedia entity is classified according to one of the four entity types in our dataset (person, location, organization, and work), based on its Wikidata class. In order to map Wikidata classes into the entity types in ENEIDE, a set of parent classes was identified for each entity type in Wikidata. Once these classes were defined, their subclasses were queried from Wikidata using the \texttt{subclass} property (\texttt{P279}). This step was performed recursively for each subclass to obtain the full hierarchy. Moreover, classes belonging to multiple entity types (e.g., person and organization) were removed. The complete list of parent classes for each type in our dataset is available in Appendix \ref{appendix:a}.

The idea of mapping Wikidata classes to four coarse-grained classes (person, location, organization, and work) derives from the fact that ENEIDE and other historical datasets for EL, such as HIPE-2020~\citep{ehrmann_overview_2020}, AjMC~\citep{ajmc_2024}, and NewsEye~\citep{newseye_2021}, perform EL on similar four classes. Therefore, adopting this design choice improves the potential generalizability of our approach to other historical domains and language varieties, despite the potential loss of information from Wikidata.

Additionally, to enrich our lookup table with temporal information from Wikidata, a set of time-related properties has been identified in the KB following the approach of~\citet{graciotti2025musical}. These properties, reported in Appendix \ref{appendix:b}, have been used to find relevant dates for the candidates in the Italian Wikipedia to use them as discriminative features during the candidate re-ranking step. Since a Wikidata entity can have multiple dates associated with it, initially the earliest, the latest, and the average dates were sampled for each entity that had more than one date. Afterwards, we measured the importance of each date for computing time intervals between the entity and the document by carrying out a permutation feature importance test, and we found the earliest date to be the most informative. Therefore, we decided to keep only the earliest date in cases of multiple dates.

The function of the lookup table is to convert each candidate retrieved from the index during the candidate retrieval step into a tuple of the following form:  $ T_{e_i} = (e_i, s_{e_i},t_{e_i}, d_{e_i},  L^2_{e_i})$ where $s_{e_i}$ is the label in Italian associated to $e_i$ in Wikidata, $t_{e_i}$ is the entity type associated to that entity in the lookup table, $d_{e_i}$ is the earliest date retrieved from the table, and $L^2_{e_i}$ is the Euclidean distance between $y_i$ and $y_{e_i}$ computed in the \textit{k}-NN search.

\subsection{Gradient-Boosted Trees Classifier}

In the candidate re-ranking step, each Wikipedia entity retrieved by the bi-encoder is evaluated by a GBTs classifier in order to predict the match probability for a given $(m, e_i)$ pair. The classifier is trained to distinguish correct entities from incorrect ones based on a set of pairwise similarity features $F_{m,e_i}$. This similarity is computed both at the candidate-level and at the set-level (i.e., including every entity retrieved by the bi-encoder). These features are carefully crafted in order to represent similarity across 4 dimensions: vector similarity, string similarity, type similarity, and time interval. While vector and string similarity scores are based on information coming from the retrieval step, type similarity and time interval are computed based on information coming from the dataset, such as the annotated entity type ($t_m$) and the date of the mention ($d_{m}$).  The complete list of features used to train the GBTs classifier is available in Table~\ref{tab8}. If an entity has no date in the lookup table, the interval between the date of the document and that of the entity is set to 0.

\begin{table*}[h]
\centering
\begin{tabular}{lll}
\toprule
\textbf{Similarity Type} & \textbf{Feature Name} & \textbf{Description} \\
\midrule
\multirow{5}{*}{\makecell[l]{Vector \\ Similarity}} 
& $L^2_{e_i}$ & $L^2$ distance between the entity and mention embeddings \\
& min & minimum $L^2$ distance in the candidates set \\
& max & maximum $L^2$ distance in the candidates set \\
& mean & mean of the $L^2$ distances in the candidates set \\
& median & median $L^2$ distance in the candidates set \\
\midrule
\multirow{2}{*}{\makecell[l]{String\\ Similarity}} 
& Levenshtein & Levenshtein distance between $s_m$ and $s_{e_i}$ \\
& Jaccard & Jaccard distance between the token sets in $s_m$ and $s_{e_i}$ \\
\midrule
\makecell[l]{Type\\ Similarity} & Type match & $1$ if $t_m=t_{e_i}$, $0$ otherwise \\
\midrule
\makecell[l]{Time \\ Interval} & $\Delta_{time}$ & Time delta computed as $d_{m}-d_{e_i}$ \\
\bottomrule
\end{tabular}
\caption{Pairwise similarity features used to train the GBTs classifier.}
\label{tab8}
\end{table*}

Given a candidate set of Wikipedia entities, the GBTs classifier is trained to predict the probability of a given $(m, e_i)$ pair to refer to the same entity based on the similarity features $F_{m,e_{i}}$. The output of the GBTs classifier is a probability score $p_{m,e_i}$ which estimates the correctness of each entity for disambiguating $m$. Optionally, the probability score can be used to filter candidates by specifying a threshold $\Delta_{NIL}$ and candidates are returned only if $p_{m,e_i} \ge \Delta_{NIL}$. If no candidate is returned, a mention is labeled as NIL. DELICATE is released as an open-source software\footnote{\href{https://github.com/sntcristian/DELICATE}{github.com/sntcristian/DELICATE}}~\citep{sntcristiandelicate}, while the trained models and KB (vector index and lookup table) are available on Hugging Face.\footnote{\href{https://huggingface.co/sntcristian/DELICATE_models}{huggingface.co/sntcristian/DELICATE\_models}}

\section{Strengthening DELICATE with LLMs}
\label{sec:delicate_llm}

In addition to our retrieval and re-ranking approach, we explore the possibility of enhancing the results of the re-ranker by prompting an instruction-tuned LLM. This variant of DELICATE couples the same bi-encoder and GBTs classifier used by DELICATE for candidate retrieval and re-ranking, but adopts a third step in which an open-source LLM is prompted to select the most appropriate candidate from the top-10 entities ranked by the classifier. More precisely, as first step, the BLINK$_{ITA}$~\citep{pozzi2023named} bi-encoder performs \textit{k}-NN search in the FAISS index of the Italian Wikipedia, returning the top-\textit{k} candidates ranked by Euclidean distance. In the second step, the GBT classifier re-ranks the entities based on vector, string, type and time similarities and returns the entities ranked by match plausibility. In the final step, a multilingual instruction-tuned open-source LLM is prompted to choose the correct entity from a candidate list including the top-10 highest scoring entities, along with their labels, entity types, and dates. Additionally, the LLM can return NIL if no candidate is considered an appropriate match for a given mention; therefore this strategy avoids the use of a $\Delta_{NIL}$ parameter to perform NIL prediction. In the experiments, we use as open-source instruction-tuned LLM \texttt{mistralai/Mistral-Small-24B-Instruct-2501}\footnote{\href{https://huggingface.co/mistralai/Mistral-Small-24B-Instruct-2501}{huggingface.co/mistralai/Mistral-Small-24B-Instruct-2501}}, due to its good performance on Italian reported on the EuroEval benchmark.\footnote{\href{https://euroeval.com/}{euroeval.com}}

The prompt comprises four components:
\begin{enumerate}
    \item A concise system instruction frames the LLM as an “\textit{information extraction system specialized in disambiguating entities within historical texts}”,
    \item A user instruction describes the candidate selection task, provides the document publication date and asks for a JSON response containing the chosen Wikipedia title and Wikidata ID,
    \item The annotated text is presented with the target mention marked by \texttt{<ENT></ENT>} tags,
    \item The list of top-\emph{10} candidates is provided as a JSON array, where each entry includes the Wikipedia title, Wikidata ID, type (person, location, organization, or work), and associated date from Wikidata.
\end{enumerate}

The complete prompt is given in Appendix \ref{appendix:c}. We compare the strategy of using a simple retrieval and re-ranking approach with the additional use of an LLM for candidate selection on different Italian historical benchmarks to quantify the trade-off between DELICATE's efficiency and the potential of incorporating LLMs.

\section{ENEIDE Dataset}
\label{sec:dataset}

\begin{figure*}
    \centering
    \includegraphics[width=0.6\textwidth]{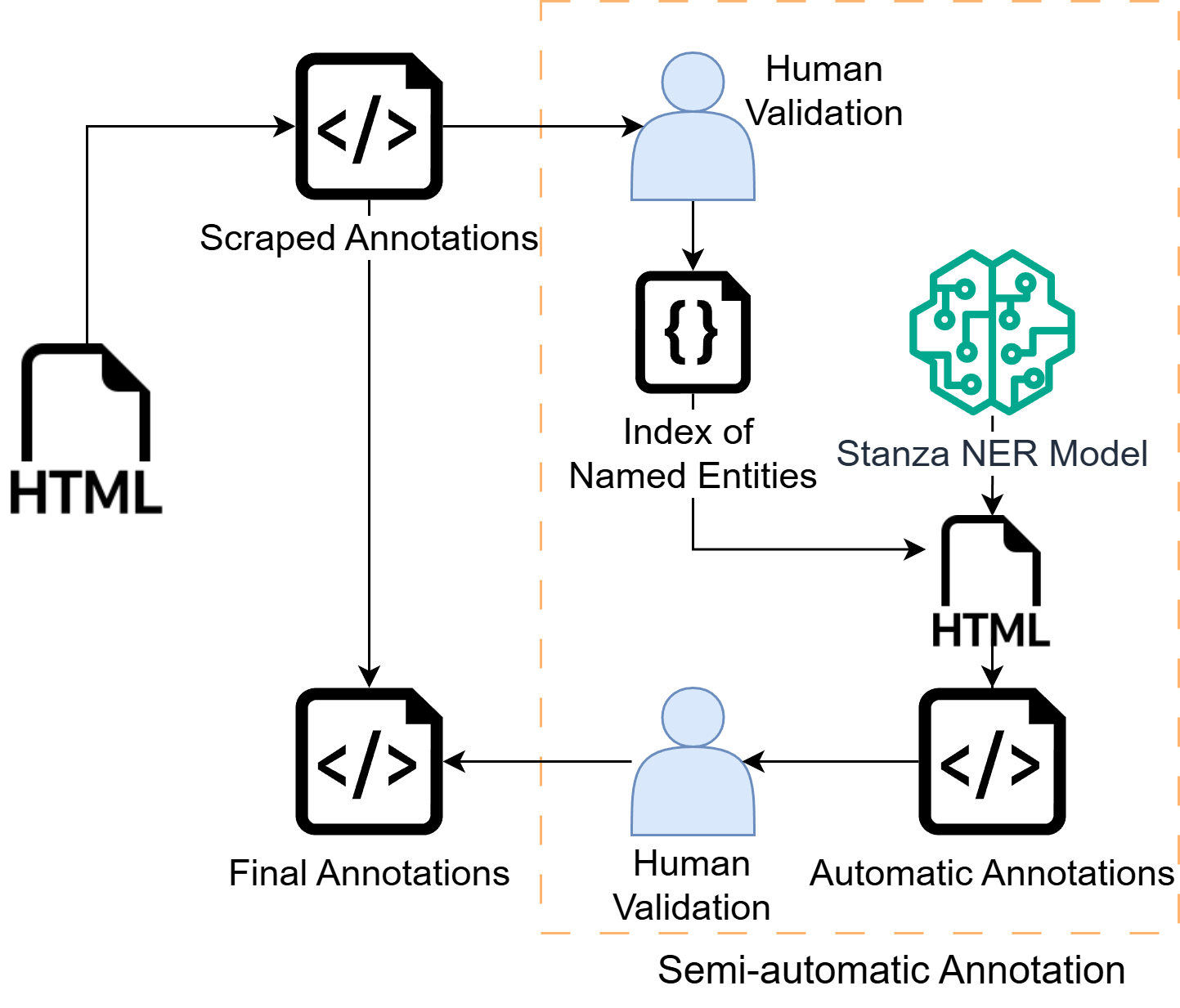}
    \caption{Data curation workflow for ENEIDE. The semi-automatic annotation step was implemented for the AMD edition to improve annotation coverage by getting additional entities not annotated in the original HTML.}
    \label{fig:eneide_wf}
\end{figure*}

The ENEIDE corpus is a diachronic, multi-domain resource for EL in Italian, semi-automatically extracted from two SDEs: \textit{Digital Zibaldone} (DZ) and \textit{Aldo Moro Digitale} (AMD). These sources span two centuries (19th to 20th) and offer heterogeneous textual genres, enabling the evaluation of EL systems on humanistic documents rich in contextual and diachronic challenges.

DZ is a TEI/XML-encoded digital edition of Giacomo Leopardi’s \textit{Zibaldone di pensieri}, a collection of over 4,500 pages of reflections on diverse intellectual domains, such as literature, history and philosophy, written between 1817 and 1832. The edition is encoded in HTML, with internal and external references including named entities linked to Wikidata when available. AMD, on the other hand, presents the complete works of the Italian politician Aldo Moro (1916–1978), including political, legal, and journalistic texts spanning from the 1930s to 1978. Encoded in RDFa, AMD uses semantic web annotations embedded in HTML to reference people, organizations, and places, aligning with Wikidata.

Together, these corpora offer a broad typology of entity references. DZ annotations cover ``person" (\texttt{PER}), ``location" (\texttt{LOC}), and ``literary work" (\texttt{WORK}) types, while AMD includes ``person", ``location", and ``organization" (\texttt{ORG}). In both, indirect references and historical variations in entity surface forms pose a challenge for EL systems. Additionally, both datasets contain entities not covered by Wikidata, introducing the task of NIL prediction, i.e., identifying entities that do not correspond to any known entry in a target KB.

\begin{table*}[h]
\centering
    \begin{tabular}{lll}
    \toprule
     \textbf{Statistic} & \textbf{DZ} & \textbf{AMD} \\
    \midrule
    Annotations present & 403 & 215 \\
    Wrong annotations & 13 & 18 \\
    Missing annotations & 10 & 64 \\
    Precision & 96.8 & 91.6 \\
    Recall & 94.4 & 70.6 \\
    F1 & 95.6 & 79.8 \\
    \bottomrule
    \end{tabular}
\caption{Quality assessment statistics for samples in DZ and AMD}
\label{tab:tab1}
\end{table*}

\begin{table*}[h]
\centering
\begin{tabular}{lccccccc}
\toprule
\textbf{Dataset} & \multicolumn{3}{c}{\textbf{Docs}} & \multicolumn{3}{c}{\textbf{Annotations}} & \makecell{\textbf{\% Overlap} \\ \textbf{(train+dev vs test)}} \\
\cmidrule(lr){2-4} \cmidrule(lr){5-7}
& \textbf{train} & \textbf{dev} & \textbf{test} & \textbf{train} & \textbf{dev} & \textbf{test} & \\
\midrule
DZ & 735 & 157 & 158 & 2935 & 727 & 617 & 93.19 \\
AMD & 743 & 159 & 160 & 2766 & 604 & 657 & 75.38 \\
\bottomrule
\end{tabular}
\caption{Statistics about number of documents, number of annotations and number of overlapping annotations in ENEIDE.}
\label{tab:tab4}
\end{table*}

\begin{table*}[h]
\centering
\begin{tabular}{lcccccccccc}
\toprule
& \multicolumn{5}{c}{\textbf{DZ}} & \multicolumn{5}{c}{\textbf{AMD}} \\
\cmidrule(lr){2-6} \cmidrule(lr){7-11}
\textbf{Split} & \textbf{PER} & \textbf{LOC} & \textbf{WORK} & \textbf{NIL} & \textbf{IDs} & \textbf{PER} & \textbf{LOC} & \textbf{ORG} & \textbf{NIL} & \textbf{IDs} \\
\midrule
train & 1661 & 488 & 786 & 182 & 623 & 759 & 940 & 1067 & 64 & 583 \\
dev & 375 & 149 & 203 & 74 & 276 & 158 & 226 & 206 & 13 & 203 \\
test & 318 & 130 & 169 & 42 & 241 & 194 & 190 & 205 & 9 & 238 \\
\bottomrule
\end{tabular}
\caption{Fine-grained statistics about number of annotations per class, number of NIL entities and unique entities (i.e. Wikidata identifiers) in each dataset split.}
\label{tab:tab5}
\end{table*}

Annotations were extracted using \texttt{Beautiful Soup}, identifying entities through HTML links. To construct the corpus, different sampling strategies were applied to each digital edition. For DZ, two subsections of the Zibaldone were extracted: one from pages 1000 to 2001 (written in 1821) and another from pages 2700 to 4000 (written in 1823). For AMD, the first paragraph of each document in the collection was sampled. To ensure quality and a balanced distribution throughout the corpus, texts with anomalous word counts relative to a normal distribution were excluded. Moreover, only documents containing at least one named entity were retained. This process resulted in a corpus of 1050 items from DZ and 1061 items from AMD. For both datasets, train, validation, and test splits were created by dividing the dataset with a 70/15/15 ratio using stratified sampling for maintaining a similar chronological distribution in all splits. Annotation quality was evaluated by domain experts using 100 documents per dataset. As shown in Table~\ref{tab:tab1}, DZ annotations scored highest with an F1 of 95.6, while AMD scored lower due to missing annotations (Recall: 70.6).

To improve AMD coverage, a semi-automatic enhancement pipeline was developed (Figure~\ref{fig:eneide_wf}). It included (i) extraction and expert validation of frequent surface forms with Wikidata IDs; (ii) tagging of missing mentions in the text; (iii) application of the Italian StanzaNLP NER model to identify unannotated entities; and (iv) final expert validation. This multi-step process increased annotation completeness without employing a fully manual annotation process.

Tables~\ref{tab:tab4} and \ref{tab:tab5} provide detailed statistics for both datasets across all splits, including annotation counts, unique entity identifiers, NIL entities, and the percentage of annotation overlap between training and test. The high overlap in DZ (93.19\%) contrasts with AMD (75.38\%), reflecting the different natures of the two corpora. The ENEIDE dataset is released on CC-BY-NC-SA license on Github\footnote{\href{https://github.com/sntcristian/ENEIDE}{github.com/sntcristian/ENEIDE}} and Zenodo \cite{sntcristianeneide}.

\section{Experimental Results}
\label{sec:experiments}

\subsection{Experimental Setup}
\label{sec:setup}

We evaluated DELICATE against other SoTA models on two EL datasets, ENEIDE and MHERCL-ITA, focusing on both ED and end-to-end EL tasks. While ENEIDE is described in Section~\ref{sec:dataset}, MHERCL-ITA was recently introduced as an Italian expansion of an English EL dataset extracted from a corpus of music periodicals~\citet{graciotti2025ke}. Differently from ENEIDE, MHERCL-ITA annotates named entities based on 34 fine-grained types. This dataset was used to test the generalizability of our approach on a different domain. In order to apply DELICATE's entity type heuristic, a manual mapping was done between the classes in ENEIDE and those in MHERCL-ITA (see Appendix~\ref{appendix:mapping}). 

In the experiments, DELICATE was compared with general and specialized baselines:

\begin{itemize}
    \item BLINK$_{ITA}$\footnote{\href{https://github.com/rpo19/pozzi_aixia_2023}{github.com/rpo19/pozzi\_aixia\_2023}}~\citep{pozzi2023named}: a transformer-based ED model for Italian, which performs dense candidate retrieval using embedding similarity. This model is also used in the candidate retrieval step of DELICATE.
    \item C-BLINK$_{ITA}$: a customized BLINK variant for historical texts introduced in~\citet{graciotti2025musical}, which filters candidates retrieved by the bi-encoder using boolean rules based on entity type matching and temporal consistency.
    \item mGENRE\footnote{\href{https://github.com/facebookresearch/GENRE}{github.com/facebookresearch/GENRE}}~\citep{de2022multilingual}: a multilingual ED system based on BERT which is trained to generate Wikipedia entities from in-text mentions in an auto-regressive way.
    \item ChatEL~\citep{ding2024chatel}: a generative approach which uses a three-step prompting framework where an LLM generates augmented context for each in-text mention and subsequently performs candidate selection from the top-10 candidates retrieved by BLINK.
\end{itemize}

Two distinct variants of the DELICATE architecture were tested, (i) DELICATE$_{GBT}$ and (ii) DELICATE$_{GBT+LLM}$. DELICATE$_{GBT}$ adopts a GBTs classifier fine-tuned on ENEIDE for candidate re-ranking and NIL prediction; while DELICATE$_{GBT+LLM}$, adopts an open-source instruction-tuned LLM which takes as input the top-10 entities scored by the GBTs reranker and performs zero-shot candidate selection and NIL prediction (as detailed in Section \ref{sec:delicate_llm}). One of the advantages of DELICATE is the small computational cost of fine-tuning, since the weights of the pre-trained BERT bi-encoder are frozen and the only component which is trained is the supervised GBTs classifier. Moreover, DELICATE$_{GBT+LLM}$, compared to other LLM baselines, such as ChatEL, has the advantage of using the generative model for a single inference, reducing drastically the cost of deploying it on large historical datasets. Moreover, the comparison with ChatEL allows to estimate the impact of incorporating information from Wikidata (e.g., entity types and dates) in the candidate selection prompt and performing candidate re-ranking. In order to make a fair comparison, both ChatEL and DELICATE$_{GBT+LLM}$ are paired with the same \texttt{mistralai/Mistral-Small-24B-Instruct-2501}.

For training the classifier, we retrieved the top-\textit{k} candidates for each mention in the training data using the BLINK bi-encoder and from each block of candidates we sampled a positive candidate $c_{pos}$, if present, and a set of negatives $C_{neg}$ containing wrong entities with $L^2$ distances evenly distributed along the median. Hyper-parameter search was carried out for 100 iterations to find the best parameters of the GBTs model for each dataset partition. The best hyper-parameters over which the results are reported are shown in Table~\ref{tab:hyper-par}. The $\Delta_{NIL}$ parameter was selected empirically by finding the best threshold on the development set of each dataset partition. For DZ, $\Delta_{NIL}$ was set to $0.4$ and for AMD it was set to $0.2$. A detailed analysis of the impact of the block size $k$ on candidate recall and the sensitivity of the $\Delta_{NIL}$ threshold for NIL prediction is provided in Appendix~\ref{appendix:d}.

While in our tests on DZ and AMD we paired DELICATE with a GBTs classifier trained independently on a single dataset partition, on MHERCL we tested the effectiveness of the re-ranker trained on the full corpus of ENEIDE.

\begin{table*}[h]
\centering
\begin{tabular}{llll}
\toprule
\textbf{Hyper-parameter} & \textbf{DZ} & \textbf{AMD} & \textbf{ENEIDE}\\
\midrule
Learning rate            & 0.115       & 0.185      & 0.135  \\
Maximum depth            & 11          & 14         & 8\\
Minimum samples leaf     & 0.0155      & 0.08       & 0.01\\
Minimum samples split    & 0.015       & 0.02       & 0.037\\
Number of estimators     & 350         & 300        & 500\\
Block size               & 50          & 20         & 50  \\
$C_{neg}$ size           & 10          & 6          & 8  \\
\bottomrule
\end{tabular}
\caption{Best hyper-parameters of the GBTs model for each dataset partition.}
\label{tab:hyper-par}
\end{table*}

On ENEIDE all models were tested in two tasks: ED and end-to-end EL. In the ED task, a model should predict the correct identifier for a sequence of tokens containing an entity mention. In the end-to-end EL task, each ED model is paired with a NER model and is evaluated on the task of identifying entity mentions within a sentence and correctly linking them. Since all the architectures tested only perform ED,  for the end-to-end EL experiments we paired all models with a GliNER model~\citep{zaratiana2023gliner} fine-tuned independently on the whole ENEIDE corpus for four epochs, given the good performances of this architecture for historical Italian reported in~\citet{santini_named_2024}.\footnote{The NER model is available on Huggingface: \href{https://huggingface.co/sntcristian/GliNER_ENEIDE}{huggingface.co/sntcristian/GliNER\_ENEIDE}} With respect to MHERCL-ITA, we evaluated the generalizability of DELICATE models on a different domain by evaluating only the ED task. 

With respect to the evaluation metrics, the ED task was assessed using accuracy, both micro and macro-averaged across all entity types. For this setting, accuracy was preferred over precision, recall and F-1 for the fact that spans fed to the bi-encoder are taken from the ground-truth annotation. However, end-to-end EL is evaluated with micro-averaged precision, recall, and F-1 since some entities may be missed or falsely recognized in the NER step. For end-to-end EL, metrics are computed in two settings: \textit{exact} and \textit{fuzzy}. Both approaches require that the predicted entity identifier matches the entity in the ground truth, but they differ in how they handle mention detection. The exact matching criterion considers a detected mention as correct only when the detected tokens perfectly align with the ground truth, while the fuzzy matching criterion allows for partial overlaps between the predicted and ground truth tokens with a minimum overlap of one token.

Finally, in order to gain insights on the explainability of DELICATE and the interpretability of its results with respect to other ED models, we carried an explainability analysis on models' features and a correlation test between entity scores and ground truth annotations. More specifically, we carried a permutation test to investigate feature relevance for classifying positive and negative candidates during re-ranking. For analysing the interpretability of DELICATE's scores, we carried a point-biserial correlation test to analyse the relation between predictions' scores of each candidate entity and their correctness. All the experiments were conducted on a Dell7920 machine equipped with an Nvidia RTX A6000 GPU.

\subsection{Evaluation on ENEIDE}
\label{sec:eneide_results}

\begin{table*}[h]
\centering
\begin{tabular}{@{}p{0.8cm}lccccr@{}}
\toprule
 & \textbf{Model} & \textbf{Acc$_{Micro}$} & \textbf{Acc$_{PER}$} & \textbf{Acc$_{LOC}$} & \textbf{Acc$_{\{WORK, ORG\}}$} & \textbf{Acc$_{Macro}$} \\
\midrule
\multirow{6}{*}{\rotatebox{90}{DZ}} 
& BLINK$_{ITA}$ & 50.24 & 56.92 & 82.30 & 13.02 & 50.75 \\
& C-BLINK$_{ITA}$ & 38.41 & 54.40 & 29.23 & 15.38 & 33.00 \\
& mGENRE & 57.86 & 67.92 & \underline{84.62} & 18.34 & 56.96 \\
& ChatEL & 62.40& 68.24 & 80.00 & \textbf{37.87} & \underline{62.04} \\
& DELICATE$_{GBT}$ & \underline{62.88}& \underline{68.87} & \textbf{86.92} & 33.14 & \textbf{62.98} \\
& DELICATE$_{GBT+LLM}$ & \textbf{63.05}& \textbf{72.96} & 73.85 & \underline{36.09} & 60.97 \\
\midrule
\multirow{6}{*}{\rotatebox{90}{AMD}} 
& BLINK$_{ITA}$ & 64.35 & 59.28 & 84.74 & 50.24 & 64.75 \\
& C-BLINK$_{ITA}$ & 49.24 & 60.82 & 36.32 & 50.24 & 49.13 \\
& mGENRE & 68.93 & 59.79 & \underline{85.79} & 61.95 & 69.18 \\
& ChatEL & 70.80 & 67.01 & 80.00 & \underline{65.85} & 70.95 \\
& DELICATE$_{GBT}$& \underline{71.47} & \underline{66.49} & \textbf{87.37} & 61.46 & \underline{71.78} \\
& DELICATE$_{GBT+LLM}$ & \textbf{72.67} & \textbf{70.62} & 78.95 & \textbf{68.78} & \textbf{72.78} \\
\bottomrule
\end{tabular}
\caption{ED results computed in micro and macro-averaged accuracy, as well as separately for all classes. For each evaluation metric, bold and underlined represent best and second best performance respectively.}
\label{tab:tab9}
\end{table*}

Table~\ref{tab:tab9} shows the results of all models performing ED on ENEIDE. For DZ, the best-performing model in terms of micro-averaged accuracy is DELICATE$_{GBT+LLM}$, with a score of 63.05, closely followed by DELICATE$_{GBT}$ (62.88) and ChatEL (62.40). DELICATE$_{GBT}$ also reports the best results in location disambiguation (86.92) and the highest macro-averaged accuracy (62.98). Overall, the most difficult class to disambiguate remains the ``work'' class, with the best-performing model, ChatEL, reaching an accuracy of 37.87, followed by DELICATE$_{GBT+LLM}$ (36.09) and DELICATE$_{GBT}$ (33.14). This result aligns with previous findings on similar datasets containing references to literary works~\citep{ehrmann2022overview}, primarily due to the high prevalence of NIL entities in this category (96.43\% of NIL entities in the test set are literary works). Notably, DELICATE$_{GBT+LLM}$ achieves the best performance in person disambiguation (72.96), suggesting that the LLM excels at leveraging contextual and temporal cues for resolving person entities, even when the overall micro-averaged accuracy remains comparable to the GBT-only variant.

In the AMD dataset, a similar trend is observed: DELICATE$_{GBT+LLM}$ achieves the best performance, with a micro-averaged accuracy of 72.67, followed by DELICATE$_{GBT}$, which scores 71.47. DELICATE$_{GBT+LLM}$ also attains the highest macro-averaged accuracy (72.78) and performs particularly well in the disambiguation of person entities (70.62), surpassing all other models. The remarkable improvement in the disambiguation of people is likely due to the relevance of temporal information in resolving these entities. For instance, entities referenced by their social role (e.g., ``the Pope''), which frequently appear in AMD, can easily lead to false positives if temporal information is not used in the disambiguation process. ChatEL achieves competitive results (70.80 micro-averaged accuracy), outperforming mGENRE and the base BLINK$_{ITA}$ bi-encoder, yet falling short of both DELICATE variants. This suggests that while LLM-based context augmentation is effective, the combination of supervised re-ranking with LLM-based candidate selection in DELICATE$_{GBT+LLM}$ yields a more robust approach.

Our experiments further reveal significant insights regarding the application of LLMs for ED across different textual domains. When comparing ChatEL with DELICATE$_{GBT}$ across both datasets, we observe that the zero-shot LLM-based approach achieves competitive performance, particularly on the ``work'' and ``organization'' classes where context augmentation plays a crucial role. In fact, these entities may exhibit a high variety in terms of surface forms and context augmentation with LLMs may be an effective solution. 

The performance variation between models prompted an investigation of candidate filtering strategies, revealing a clear disparity between methods. The rule-based constraint methodology employed in C-BLINK$_{ITA}$ for filtering bi-encoder outputs proves markedly less effective than supervised re-ranking techniques. Systematic error analysis indicates that many misclassification errors originate from the inappropriate inclusion of temporally irrelevant information. For example, the Wikidata entity \textit{Italy} (Q38) records June 18, 1946, as its inception date in the KB; a historically accurate marker for the Italian Republic, yet misleading when disambiguating references to Italy in earlier historical texts.

\begin{table*}[h]
\centering
\begin{tabular}{@{}p{0.8cm}lcccccc@{}}
\toprule
& & \multicolumn{3}{c}{\textit{Exact}} & \multicolumn{3}{c}{\textit{Fuzzy}} \\
\cmidrule(lr){3-5} \cmidrule(lr){6-8}
& \textbf{Model} & \textbf{Precision} & \textbf{Recall} & \textbf{F-1} & \textbf{Precision} & \textbf{Recall} & \textbf{F-1} \\
\midrule
\multirow{6}{*}{\rotatebox{90}{DZ}} 
& LLaMA 3.1-8B & 47.12 & 39.71 & 43.10 & 47.50 & 40.03 & 43.45 \\
& C-BLINK$_{ITA}$ & 37.21 & 31.60 & 34.18 & 37.40 & 37.77 & 34.36 \\
& mGENRE & 54.62 & 46.02 & 49.96 & 55.20 & 46.50 & 50.48 \\
& ChatEL & 55.92 & 47.49 & 51.36 & 57.25 & 48.62 & 52.59 \\
& DELICATE$_{GBT}$ & \underline{56.49} & \underline{47.97} & \underline{51.88} & \underline{57.44} & \underline{48.78} & \underline{52.76} \\
& DELICATE$_{GBT+LLM}$ & \textbf{56.87} & \textbf{48.30} & \textbf{52.23} & \textbf{57.82} & \textbf{49.11} & \textbf{53.11} \\
\midrule
\multirow{6}{*}{\rotatebox{90}{AMD}} 
& BLINK$_{ITA}$ & 61.21 & 60.27 & 60.74 & 61.21 & 60.27 & 60.74 \\
& C-BLINK$_{ITA}$ & 47.65 & 46.52 & 47.08 & 47.83 & 46.69 & 47.25 \\
& mGENRE & 64.83 & 63.84 & 64.33 & 64.83 & 63.84 & 64.33 \\
& ChatEL & 66.78 & 65.20 & 65.98 & 67.13 & 65.53 & 66.32 \\
& DELICATE$_{GBT}$ & \underline{68.87} & \underline{67.23} & \underline{68.04} & \underline{69.04} & \underline{67.40} & \underline{68.21} \\
& DELICATE$_{GBT+LLM}$ & \textbf{69.39} & \textbf{67.74} & \textbf{68.56} & \textbf{69.73} & \textbf{68.08} & \textbf{68.90} \\
\bottomrule
\end{tabular}
\caption{End-to-end Entity Linking results.}
\label{tab:tab11}
\end{table*}

Table~\ref{tab:tab11} reports the performance of the evaluated models in an end-to-end EL setting, pairing all models with GliNER. In both DZ and AMD datasets, DELICATE variants outperform the baseline systems. For DZ, DELICATE$_{GBT+LLM}$ achieves the best F-1 score of 52.23 under exact matching and 53.11 under fuzzy matching, followed closely by DELICATE$_{GBT}$, which scores 51.88 and 52.76 in the exact and fuzzy settings, respectively.

For AMD, DELICATE$_{GBT+LLM}$ achieves the best results, with an F-1 score of 68.56 under exact matching and 68.90 under fuzzy matching, while DELICATE$_{GBT}$ follows closely with 68.04 and 68.21, respectively. ChatEL ranks third with 65.98 and 66.32, substantially outperforming mGENRE (64.33) but remaining below both DELICATE variants. The improvement in F-1 against the baseline systems is particularly significant, with DELICATE$_{GBT}$ outperforming even architectures with a higher number of parameters. These gains illustrate that DELICATE is not only effective at disambiguation but also robust to potential errors when integrated with a NER component, such as boundary misdetection and entity type misclassification. Moreover, the consistent improvement of DELICATE$_{GBT+LLM}$ over DELICATE$_{GBT}$ in both ED and EL suggests that the additional LLM-based candidate selection step helps to recover from errors introduced by the GBTs re-ranker, providing a further layer of contextual validation.

\subsection{Evaluation on MHERCL-ITA}
\label{sec:mhercl_experiments}

The generalizability of DELICATE models is tested on MHERCL-ITA, a public available EL evaluation dataset extracted from Italian music periodicals published in the 20th century. It is to be noted that MHERCL-ITA presents several challenges which are relevant for historical EL: (i) texts often present OCR errors, which may cause problems for models trained on web-crawled texts, (ii) MHERCL exhibits a higher concentration of long-tail entities (entities with low popularity in Wikipedia/Wikidata) if compared with general-purpose datasets, (iii) this dataset also presents a higher percentage of NIL entities if compared with the ENEIDE test set (28.67\% vs 6.39\%). In the experiments we applied pre-trained DELICATE$_{GBT}$ models following the setup used for DZ, i.e. using $50$ as block size and $0.4$ as $\Delta_{NIL}$. For DELICATE$_{GBT+LLM}$, the top-10 entities ranked by the GBTs classifier trained on the full ENEIDE corpus are passed to the LLM for candidate selection and NIL prediction with the prompt detailed in Appendix~\ref{appendix:c}.

\begin{table*}[h]
\centering
\begin{tabular}{lcc}
\toprule
\textbf{Model} & \textbf{Acc$_{Micro}$} & \textbf{Acc$_{Macro}$} \\
\midrule
BLINK$_{ITA}$ & 41.69 & 39.97 \\
C-BLINK$_{ITA}$ & 31.70 & 28.16 \\
mGENRE & 45.84 & 44.95 \\
ChatEL & \underline{69.62} & \underline{68.86} \\
DELICATE$_{GBT}$ & 58.78 & 56.97 \\
DELICATE$_{GBT+LLM}$ & \textbf{70.44} & \textbf{69.49} \\
\bottomrule
\end{tabular}
\caption{ED results of DELICATE and baseline models on MHERCL-ITA.}
\label{tab:merchl_results}
\end{table*}

Table~\ref{tab:merchl_results} reports micro- and macro-averaged accuracy for all models on the ED task of MHERCL-ITA. The results on this out-of-domain dataset reveal a markedly different pattern compared to ENEIDE, with LLM-based approaches achieving substantially stronger performance. DELICATE$_{GBT+LLM}$ obtains the best results overall, with a micro-averaged accuracy of 70.44\% and a macro-averaged accuracy of 69.49\%, followed closely by ChatEL (micro: 69.62\%, macro: 68.86\%). Both LLM-enhanced models substantially outperform the GBT-only variant, DELICATE$_{GBT}$, which achieves 58.78\% micro-averaged and 56.97\% macro-averaged accuracy. The performance gap of approximately 12 percentage points between DELICATE$_{GBT+LLM}$ and DELICATE$_{GBT}$ can be largely attributed to the higher proportion of NIL entities in MHERCL-ITA (28.67\%), which the LLM handles more effectively through its ability to assess whether any candidate is a plausible match for a given mention, without relying on a fixed $\Delta_{NIL}$ threshold.

These results highlight a key finding: while DELICATE$_{GBT}$ alone demonstrates strong cross-domain generalizability compared to non-LLM baselines, the addition of an LLM in the candidate selection step proves particularly beneficial in domains with high NIL rates and long-tail entities. The combination of supervised re-ranking with LLM-based candidate selection in DELICATE$_{GBT+LLM}$ yields the most robust approach overall, leveraging the GBTs classifier to produce a well-ranked candidate list while exploiting the LLM's contextual reasoning capabilities for the final selection. Moreover, the fact that DELICATE$_{GBT+LLM}$ slightly outperforms ChatEL suggests that the supervised re-ranking step provides a useful inductive bias even when an LLM is used downstream, by presenting the model with a more informative and better-ordered set of candidates.

\subsection{Explainability Analysis}
\label{sec:statistics}

\begin{figure*}
    \centering
    \includegraphics[width=1\textwidth]{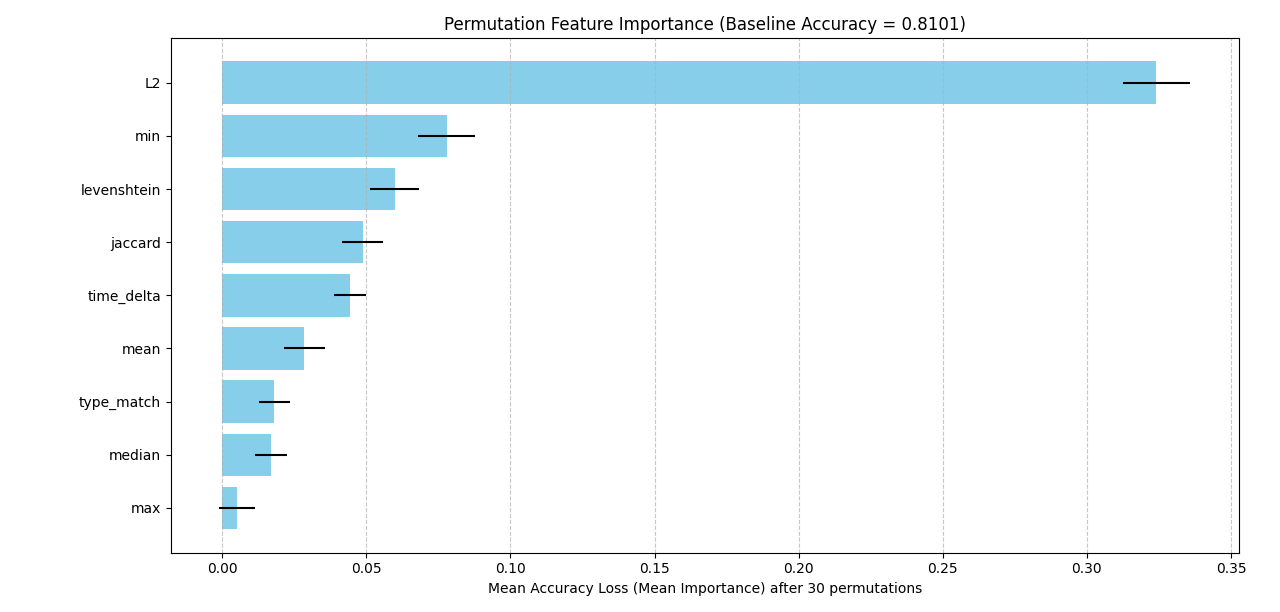}
    \caption{Plot of mean importance of each feature of the GBTs model after 30 Permutation Tests on the full ENEIDE test set.}
    \label{fig:features}
\end{figure*}

Due to the importance of explainability and interpretability in NLP applications for the humanities, we further analysed the performance of DELICATE by adopting two standard statistical tests. More specifically, a permutation test was carried out on feature relevance in order to estimate the importance of each feature in leading the GBTs component to classify positive and negative candidates. Moreover, the point-biserial coefficient was used to analyse the degree of correlation between prediction scores of each model and the rate of true and false positives predicted in the ED task.
The plots of the mean accuracy losses obtained after 30 permutation tests for DELICATE$_{GBT}$ are shown in Figure~\ref{fig:features}. As shown, vector-based similarity distances are the most important for the GBTs model, closely followed by string similarity ones, i.e., Levenshtein and Jaccard distances. Moreover, permutation tests show how time intervals have high relevance, being the fifth most important feature with a mean importance of $\approx$ 0.05. This suggests the relevance of temporal constraints in reducing the rate of false positives in historical texts.

\begin{table*}[h]
\centering
\begin{tabular}{llcc}
\toprule
 & \textbf{Model} & \textbf{Point-biserial Correlation} & \textbf{p-value} \\
\midrule
\multirow{6}{*}{\rotatebox{90}{DZ}} & BLINK$_{ITA}$ & 0.5986 & 2.9675e-61 \\
& C-BLINK$_{ITA}$ & 0.4270 & 9.2751e-29 \\
& mGENRE & 0.5969 & 7.8370e-61 \\
& ChatEL & 0.5142 & 1.4821e-43 \\
& DELICATE$_{GBT}$ & \textbf{0.7973} & 5.4713e-137 \\
& DELICATE$_{GBT+LLM}$ & 0.5587 & 8.3192e-51 \\
\midrule
\multirow{6}{*}{\rotatebox{90}{AMD}} & BLINK$_{ITA}$ & 0.1490 & 2.8596e-04 \\
& C-BLINK$_{ITA}$ & 0.1413 & 5.8373e-04 \\
& mGENRE & 0.4287 & 1.0010e-27 \\
& ChatEL & 0.3054 & 4.1263e-14 \\
& DELICATE$_{GBT}$ & \textbf{0.6848} & 9.8293e-83 \\
& DELICATE$_{GBT+LLM}$ & 0.3391 & 2.7145e-17 \\
\bottomrule
\end{tabular}
\caption{Point-biserial Correlation between prediction scores and correct predictions for all ED models. High scores are an index of a strong correlation between prediction scores and correctness of the result.}
\label{tab:tab12}
\end{table*}

Table~\ref{tab:tab12} presents the point-biserial correlation between prediction scores and correct predictions for all ED models on ENEIDE. The strongest correlations are observed for DELICATE$_{GBT}$, which achieves 0.7973 on DZ and 0.6848 on AMD, indicating that the GBTs confidence scores are strongly associated with correct disambiguation decisions. This suggests that supervised methods, such as GBTs, despite having the disadvantage of requiring training data, produce more interpretable scores than the approaches based on language models, such as BLINK and mGENRE.

Notably, DELICATE$_{GBT+LLM}$ and ChatEL exhibit substantially lower point-biserial correlations (0.5587 and 0.5142 on DZ; 0.3391 and 0.3054 on AMD, respectively) despite achieving competitive or superior accuracy on certain metrics. This suggests that, while incorporating an LLM can improve raw accuracy (particularly for NIL prediction), it may reduce the interpretability of the system's confidence measures.

This finding is relevant with respect to the black box issue that language models usually have, since it demonstrates how it can be tackled by employing more traditional ML methods in downstream tasks. The degree of interpretability offered by the GBT-only variant is particularly valuable for applications that require reliable confidence measures and suggests the importance of integrating DELICATE in human-in-the-loop systems, where it could be paired with a domain expert for verifying and consolidating the results. However, investigating this application of DELICATE goes beyond the scope of this study.

\section{Discussion and Conclusion}

\label{sec:discussion}

\paragraph{Discussion.}
The experimental results indicate that the proposed DELICATE architecture offers several notable advantages while also revealing challenges that require further investigation. First of all, DELICATE achieves significant efficiency gains by fine-tuning only the supervised GBTs and keeping the BERT bi-encoder weights frozen. This design choice allows for rapid adaptation even on CPU-based systems, which is especially beneficial for settings with limited computational resources or when training samples are few.

Moreover, the system shows a reduced bias towards under-represented entities, as highlighted by improved performance on long-tail entities in ENEIDE and MHERCL-ITA. This improvement suggests that DELICATE’s candidate re-ranking mechanism is more resilient to domain shifts and chronological variations. Additionally, by integrating type and time information from Wikidata, the model enhances its disambiguation capabilities in historical documents where contextual information is often necessary.

A further evidence of DELICATE’s generalizability is its strong performance on the out-of-domain MHERCL-ITA dataset. Despite being fine-tuned solely on ENEIDE, DELICATE$_{GBT+LLM}$ achieves micro- and macro- averaged accuracies around 70\%, substantially outperforming both generic baselines and more complex architectures employing bi-encoders paired with LLMs.

Another noteworthy strength is the adoption of GBTs in the final classification stage in DELICATE$_{GBT}$, which contributes to the explainability and interpretability of system's results. This tree-based approach not only facilitates insights into the relevance of different similarity dimensions but also yields confidence scores that correlate strongly with linking accuracy. Such characteristics are particularly valuable in historical EL pipelines, where manual verification of uncertain links is often required.

\paragraph{Limitations.} Despite these strengths, the experiments also revealed some limitations. DELICATE relies on annotated training data to fine-tune the GBTs classifier effectively, which poses a challenge when labeled resources are scarce. In addition, inaccuracies in the initial candidate retrieval step (e.g., due to OCR errors) can propagate through the final re-ranking, potentially reducing overall accuracy. Finally, the need to tune hyper-parameters (e.g., learning rate, tree depth, $\Delta_{NIL}$) for each dataset partition introduces additional complexity which however seems to not impact scalability and generalizability.

Moreover, the fact that DELICATE uses four coarse-grained classes for computing the \textit{Type match} feature may introduce some noise due to asymmetries between the entity type annotations in ENEIDE and the fine-grained classes in Wikidata. In spite of these limitations, the use of four coarse-grained classes allows DELICATE to generalise to datasets that have a similar entity type taxonomy, such as HIPE-2020, AjMC, and NewsEye, as already discussed in Section \ref{sec:lookup_table}.

Finally, treating missing dates as neutral (i.e., setting the time interval to 0) may bias disambiguation results towards entities lacking temporal information in KBs. However, this design choice was made to avoid penalising entities that lack temporal information in Wikidata, which is common for long-tail entities that are prevalent in historical texts.

\paragraph{Summary \& Conclusion.} To summarise, this paper presented DELICATE, a novel supervised method that combines a BERT-based bi-encoder and structured information from KBs to perform ED on historical documents from multiple domains and written across different chronological periods. Experimental results show that DELICATE not only outperforms general-purpose ED systems when trained with domain-specific data, but also demonstrates strong cross-domain adaptability by achieving good performances on new datasets without further fine-tuning. A further contribution of this study is the release of a novel Italian corpus for training and evaluating EL systems, called ENEIDE, which is semi-automatically extracted from two digital editions. All resources produced in this work, including datasets, source code, trained models and results, are released in open-source to enhance the reproducibility of the study. Future work may extend DELICATE to multilingual historical EL~\citep{santini-etal-2026-confidence} by refining its KB using more structured taxonomies such as YAGO-4.5~\citep{SuchanekABCPS24}. The ENEIDE corpus could also be expanded with multilingual resources to serve as a benchmark for multilingual EL systems in the humanities. Another promising direction involves the integration of DELICATE into KE pipelines for different historical documents, such as letters~\citep{santini2024combining} and books~\citep{santini_knowledge_2022}.

\newpage

\section*{Appendix A: Entity Type Mapping with Wikidata Classes}
\label{appendix:a}

To categorize entities in the Italian Wikipedia according to the four entity types used in the ENEIDE benchmark — person, organization, location, and work — a mapping between Wikidata classes and ENEIDE types was constructed. This mapping relies on a manually curated list of root Wikidata classes, chosen for each ENEIDE entity type based on their semantic alignment.

For each of these root classes, the entire subclass hierarchy was obtained using the \texttt{subclass of} property (\texttt{P279}). This ensures comprehensive coverage of Wikidata's ontological structure. Classes that appeared under more than one ENEIDE category were excluded to maintain mutually exclusive type assignments.

The resulting class-to-type mappings were used to annotate each entity in the dense index by checking whether its Wikidata class belonged to the subtree of a parent class. The root classes for each entity type used in our mapping are shown in Table~\ref{tab:tab6}.

\begin{table*}[h]
\centering
\begin{tabular}{p{3cm}p{8cm}}
\toprule
\textbf{Entity Type} & \textbf{Parent Wikidata Classes} \\
\midrule
Person & person (Q215627), fictional character (Q95074), fictional person (Q97498056) \\
Organization & governing body (Q895526), group of humans (Q16334295), fictional organization (Q14623646) \\
Location & fictional location (Q3895768), geographic entity (Q27096213), spatial object (Q58416391) \\
Work & creative work (Q17537576), intellectual work (Q15621286) \\
\bottomrule
\end{tabular}
\caption{Root Wikidata classes for different entity types in our dataset}
\label{tab:tab6}
\end{table*}

\section*{Appendix B: Wikidata time-related properties used for candidate re-ranking}
\label{appendix:b}

To enrich the knowledge base with temporal information, a curated list of time-related properties from Wikidata was created. These properties provide semantically relevant dates linked to entities, such as their inception, publication, or debut. The inclusion of temporal information allows DELICATE system to prioritize or disambiguate candidates during re-ranking, especially when the context includes temporal cues.

The list of used properties presented in Table~\ref{tab:tab7} based on the taxonomy presented in~\citet{graciotti2025musical}, and each property was selected for its relevance across the four ENEIDE entity types.

\begin{table*}[h]
\centering
\begin{tabular}{ll}
\toprule
\textbf{Property} & \textbf{Name} \\
\midrule
P569 & Date of Birth \\
P571 & Inception \\
P1619 & Date of Official Opening \\
P1191 & Date of First Performance \\
P10135 & Recording Date \\
P577 & Publication Date \\
P575 & Time of Discovery or Invention \\
P1317 & Floruit \\
P7124 & Date of the First One \\
P10673 & Debut Date \\
P9448 & Introduced On \\
P6949 & Announcement Date \\
P729 & Service Entry \\
P2031 & Work Period (Start) \\
P585 & Point in Time \\
\bottomrule
\end{tabular}
\caption{Time-related Wikidata properties used in the lookup table.}
\label{tab:tab7}
\end{table*}

\newpage

\section*{Appendix C: DELICATE$_{GBT+LLM}$ Prompt Template}
\label{appendix:c}

This appendix presents the full prompt used in DELICATE$_{GBT+LLM}$ to perform entity disambiguation, translated in English for increased readability. The prompt includes a task instruction, the input text with the target mention marked by \texttt{<ENT></ENT>} tags, and a JSON block listing candidate entities with associated type and temporal information.

\begin{quote}
\textbf{System Prompt:}
You are an effective information extraction system specialized in disambiguating entities within historical texts.
Your task is to analyse the text provided by the user and disambiguate the reference marked by <ENT></ENT> tags by selecting a Wikidata entity from a given list of candidates.
Always respond by returning a JSON-formatted answer; do not generate Python code.
\end{quote}

\begin{quote}
\textbf{User Prompt:}
Read the input text published in \texttt{\{document\_date\}}.

Disambiguate the entity mentioned between the <ENT></ENT> tags by selecting the most appropriate Wikidata entity from the list of candidates.

Return the corresponding Wikipedia title and Wikidata ID of the selected entity in a JSON object formatted as follows:
\texttt{
\{
"wikipedia\_title": "", "wikidata\_id": ""
\}
}

Make sure to select both the Wikipedia title and the Wikidata ID from the provided list of candidates.

If none of the candidates match the entity tagged with <ENT></ENT>, return an empty JSON object.

Input Text: \texttt{\{annotated\_text\}}

List of Candidates: \texttt{\{candidates\_in\_json\}}
\end{quote}

\section*{Appendix D: Entity Type Mapping from MHERCL-ITA to ENEIDE}
\label{appendix:mapping}

To enable the application of DELICATE’s entity type heuristic on MHERCL-ITA, we manually mapped the 34 fine-grained entity types defined in MHERCL-ITA to the four coarse-grained categories used in ENEIDE: \texttt{PER} (person), \texttt{ORG} (organization), \texttt{LOC} (location), and \texttt{WORK} (work). Table~\ref{tab:type-mapping} summarizes this mapping.

\begin{table*}[h]
\centering
\begin{tabular}{p{3cm}p{8cm}}
\toprule
\textbf{ENEIDE Type} & \textbf{MHERCL-ITA Fine-Grained Types} \\
\midrule
Person & person \\
Organization & family, organization, school, government-organization, university, newspaper, magazine \\
Location & city, country, country region, continent, location, mountain, road, lake, island, building, worship-place, facility, theater \\
Work & book, work-of-art, publication, music, music key, award, event, festival, court decision, war, conference, law \\
\bottomrule
\end{tabular}
\caption{Manual mapping of MHERCL-ITA fine-grained types to ENEIDE's coarse-grained types used by DELICATE.}
\label{tab:type-mapping}
\end{table*}

\section*{Appendix E: Block Size and NIL Threshold Analysis}
\label{appendix:d}

This appendix provides empirical justification for two key hyper-parameters of DELICATE$_{GBT}$: the block size $k$ used in the $k$-NN candidate retrieval step and the NIL threshold $\Delta_{NIL}$ used to filter low-confidence predictions. Both analyses are conducted on the development sets of DZ and AMD to ensure that the final parameter choices are grounded in systematic evaluation rather than ad hoc tuning.

\subsection*{Impact of Block Size on Candidate Recall}

Since DELICATE's re-ranking step can only select among the candidates retrieved by the bi-encoder, the quality of the initial retrieval directly bounds the overall system performance. Table~\ref{tab:recall_k} reports Recall@$k$ at increasing block sizes on the development sets of DZ and AMD, measuring the proportion of mentions for which the correct entity appears within the top-$k$ candidates returned by BLINK$_{ITA}$.

\begin{table*}[h]
\centering
\begin{tabular}{lcc}
\toprule
\textbf{Block Size ($k$)} & \textbf{DZ} & \textbf{AMD} \\
\midrule
1  & 0.539 & 0.653 \\
3  & 0.603 & 0.720 \\
5  & 0.628 & 0.757 \\
10 & 0.652 & 0.788 \\
20 & 0.682 & 0.822 \\
30 & 0.692 & 0.839 \\
40 & 0.697 & 0.852 \\
50 & 0.699 & 0.864 \\
\bottomrule
\end{tabular}
\caption{Recall@$k$ of the BLINK$_{ITA}$ bi-encoder at different block sizes on the development sets of DZ and AMD.}
\label{tab:recall_k}
\end{table*}

For both datasets, recall increases monotonically with $k$, but with diminishing returns at larger block sizes. On AMD, recall grows substantially from $k{=}1$ (0.653) to $k{=}20$ (0.822), after which the marginal gain per additional candidate decreases noticeably (e.g., only +0.042 from $k{=}20$ to $k{=}50$). This motivated the selection of $k{=}20$ as the block size for training the GBTs classifier on AMD, as it offers a favourable trade-off between recall coverage and the computational cost of re-ranking a larger candidate set.

On DZ, recall values are consistently lower across all block sizes, reflecting the higher prevalence of long-tail entities in the literary domain that are poorly represented in the bi-encoder's training data. Even at $k{=}50$, recall reaches only 0.699, compared to 0.864 for AMD. This lower retrieval ceiling motivated the choice of a larger block size ($k{=}50$) for DZ, in order to maximise the likelihood of including the correct entity in the candidate set before re-ranking.

\subsection*{NIL Threshold Sensitivity}

Table~\ref{tab:nil_sensitivity} reports the performance of NIL prediction at different values of $\Delta_{NIL}$ on the development sets of DZ and AMD, measured in terms of precision, recall, and F-1 over the subset of NIL entities. This analysis was conducted to identify the threshold that yields the best balance between correctly identifying NIL mentions and avoiding the incorrect rejection of valid candidates.

\begin{table*}[h]
\centering
\begin{tabular}{ccccccc}
\toprule
 & \multicolumn{3}{c}{\textbf{DZ}} & \multicolumn{3}{c}{\textbf{AMD}} \\
\cmidrule(lr){2-4} \cmidrule(lr){5-7}
$\Delta_{NIL}$ & \textbf{Precision} & \textbf{Recall} & \textbf{F-1} & \textbf{Precision} & \textbf{Recall} & \textbf{F-1} \\
\midrule
0.1 & 0.208 & 0.263 & 0.233 & 0.125 & 0.111 & 0.117 \\
0.2 & 0.174 & 0.315 & 0.224 & 0.120 & 0.333 & \textbf{0.176} \\
0.3 & 0.172 & 0.395 & 0.240 & 0.112 & 0.333 & 0.166 \\
0.4 & 0.208 & 0.578 & \textbf{0.306} & 0.088 & 0.444 & 0.148 \\
0.5 & 0.172 & 0.605 & 0.289 & 0.071 & 0.444 & 0.123 \\
0.6 & 0.164 & 0.658 & 0.263 & 0.063 & 0.444 & 0.111 \\
0.7 & 0.173 & 0.763 & 0.282 & 0.067 & 0.666 & 0.122 \\
0.8 & 0.170 & 0.894 & 0.286 & 0.060 & 0.777 & 0.112 \\
0.9 & 0.148 & 0.895 & 0.255 & 0.043 & 0.777 & 0.081 \\
\bottomrule
\end{tabular}
\caption{NIL prediction performance (Precision, Recall, F-1) on the development sets of DZ and AMD at different $\Delta_{NIL}$ thresholds.}
\label{tab:nil_sensitivity}
\end{table*}

The two datasets exhibit markedly different optimal thresholds, which reflects their different NIL entity distributions. For DZ, the best F-1 (0.306) is achieved at $\Delta_{NIL}{=}0.4$. This relatively high threshold is consistent with the larger proportion of NIL entities in DZ (see Table~\ref{tab:tab5}): a more aggressive filtering strategy is needed to correctly reject the substantial number of mentions that lack a corresponding Wikidata entry, even at the cost of moderate precision. For AMD, the best F-1 (0.176) is obtained at $\Delta_{NIL}{=}0.2$, reflecting the much smaller proportion of NIL entities in this dataset. Here, a lower threshold suffices to capture the few NIL cases without excessively penalising valid candidates.

Overall, NIL prediction remains a challenging subtask for DELICATE$_{GBT}$, with relatively low F-1 scores across both datasets. This is one of the motivations behind the introduction of DELICATE$_{GBT+LLM}$ (Section~\ref{sec:delicate_llm}), which delegates NIL prediction to the LLM rather than relying on a fixed threshold, yielding substantial improvements on datasets with high NIL rates such as MHERCL-ITA (see Section~\ref{sec:mhercl_experiments}).

\section*{Declarations}

\paragraph{Data Availability} Source code, dataset and trained models produced in this research are publicly available:
\begin{itemize}
    \item DELICATE (source code): \href{https://github.com/sntcristian/DELICATE}{https://github.com/sntcristian/DELICATE}
    \item DELICATE (trained models): \href{http://doi.org/10.57967/hf/6984}{http://doi.org/10.57967/hf/6984}
    \item ENEIDE (dataset): \href{https://github.com/sntcristian/ENEIDE}{https://github.com/sntcristian/ENEIDE}
\end{itemize}

\bibliography{sn-bibliography}

\end{document}